\documentclass{article} % For LaTeX2e
\usepackage{nips15submit_e,times}
\usepackage{hyperref}
\usepackage{url}
\usepackage{graphicx}
\usepackage{amsmath}
\usepackage{amssymb}
\usepackage{subfig}

\title{Finding Optimal Combination of Kernels using Genetic Programming}
\author{
Jyothi Korra\\
Computer Science and Engineering\\
Christ University Faculty of Engineering\\
Bangalore, India\\
\texttt{jyothi.korra@christuniversity.in} \\
}

\nipsfinalcopy % Uncomment for camera-ready version
\begin{document}
\maketitle

\begin{abstract}
In Computer Vision, problem of identifying or classifying the objects present in an image is called Object Categorization. It is a challenging problem, especially when the images have clutter background, occlusions or different lighting conditions. Many vision features have been proposed which aid object categorization even in such adverse conditions. Past research has shown that, employing multiple features rather than any single features leads to better recognition. Multiple Kernel Learning (MKL) framework has been developed for learning an optimal combination of features for object categorization. Existing MKL methods use linear combination of base kernels which may not be optimal for object categorization. Real-world object categorization may need to consider complex combination of kernels(non-linear) and not only linear combination. Evolving non-linear functions of base kernels using Genetic Programming is proposed in this report. Experiment results show that non-kernel generated using genetic programming gives good accuracy as compared to linear combination of kernels. 
\end{abstract}

\section{Introduction}
Object Categorization is the problem of categorizing the given image into predefined classes of objects like face, car, bike etc. This problem is well-studied in the field of Computer Vision. Face recognition is a less harder problem than Object Categorization. There are systems where computer can identify expressions and morph faces automatically ~\cite{dg6}. Object Categorization is a challenging problem, especially when the images have clutter background, occlusions or different lighting conditions. When a computer vision system or computer vision algorithm is designed the choice of feature representation can be a critical issue. In some cases, a higher level of detail in the description of a feature may be necessary for solving the problem, but this comes at the cost of having to deal with more data and more demanding processing. In this report, an instance of a feature representation is referred to as a (feature) descriptor.  \\

In some applications like object categorization it is not sufficient to extract only one type of feature to obtain the relevant information from the image data. Instead many feature descriptors have been proposed which aid object categorization even in those adverse conditions. Each descriptor has its own merits and de-merits. Some descriptors are invariant to transformations while the others are more discriminative. Past research has shown that employing multiple feature descriptors rather than any single descriptor leads to better recognition. This report focuses on the problem of learning the optimal combination of the available descriptors for a particular classification task. Labels for classification task are mostly obtained by manual labelling process or through crowd-sourcing ~\cite{dg4}

In~\cite{kumar07,varma07,dg1, dg2}, the authors employ the Multiple Kernel Learning (MKL) framework to find the optimal combination of feature descriptors (kernels). The goal of MKL is to simultaneously optimize the combination of kernels and the usual classification objective. Existing MKL methods for combining kernels are linear combinations of base kernels (see figure 1). But non-linear combination of base kernels may be helpful to boost performance of the classifier. This would be ideal for applications such as object categorization, in which a combination of the descriptors is known to perform better than any single descriptor. 

The new hybrid model which uses Genetic Programming, for evolving kernel from base kernels and Support Vector Machine (SVM), for finding classifier function using evolved kernel from GP is proposed. This is the first attempt to have non-linear kernel combination for object categorization. Experiments will be conducted on datasets like Caltech-5, Caltech-101 etc to verify advantage of this hybrid model for improving object categorization. 
 
\section{Organization of this manual}
The outline of the report is as follows: section~\ref{sec:past} discusses the past work in this area. section~\ref{sec:gp} briefly reviews genetic programming and its advantages. section~\ref{sec:obj} describes the methodology for carrying out object
categorization using genetic programming using non-linear kernels. Section~\ref{sec:exp} gives the implementation details and results on two datasets. This is followed by the conclusion.

%%%%%%%%%%%%%%%%%%%%%%%%%%%%%%%%%%%%%%%%%%%%%%%%%%%%%%%%%%%%%%%%%%%%%%

\section{Past Work}\label{sec:past}
There have been some work in genetic programming for evolving kernels for Support Vector Machine ~\cite{1277332, 1277292, genetickernel}. \\
~\cite{genetickernel} uses Genetic Programming for evolving the kernel for SVM classifier. The approach presented there combines the two techniques of SVMs and GP, using the GP to evolve a kernel for a SVM. The goal there is to eliminate the need for testing various kernels and their parameter settings. They claim the approach might also be possible to discover new kernels that are particularly useful for the type of data under analysis. They show that their method performs better than manual choosing of the kernel and adjusting parameters. \\
~\cite{1277292} uses a set of standard kernels for evolving expression for new kernel which performs better for given problem using genetic programming. Terminal set contains feature vectors, first level from terminals in the GP trees contains only standard kernels defined before-hand. Variable set contains functions which take two kernels as arguments and provides a kernel as output. \\
~\cite{1277332} tries to learn a regression function where kernels act as the regression variables. Each GP chromosome gives the complex combination of the set of kernels that is defined already. This is closely related to ~\cite{1277332}, where they try to evolve regression function using GP where kernels acts as regression variables. This report studies how these non-linear function of base kernels in the case descriptors affect the performance of the object categorization and tries to find a way to reduce the time taken for evolving functions using GP. \\
Even though these try to evolve kernels, no work has been done on evolving non-linear kernels for object categorization. The state-of-the-art work in object categorization considers many descriptors and try to find the optimal combination of the descriptors. ~\cite{kumar07, varma07, dg1, dg2, dg3} considers combining descriptors using multiple kernel learning. Descriptors are extracted from the image and each descriptors will have many kernels formed using the feature vector of the descriptors. And these kernels are combined using the Multiple Kernel Learning(MKL) in Support Vector Machine(SVM) framework. The principle idea is to combine kernels linearly. But real-world object categorization may perform better when we have non-linear combination of these descriptors. In our work we find non-linear kernel combination for improving performance. The next section explains genetic programming and how to evolve non-linear kernel combination using GP.

\begin{figure}
\centering
\includegraphics[width=3in,height=3.5in]{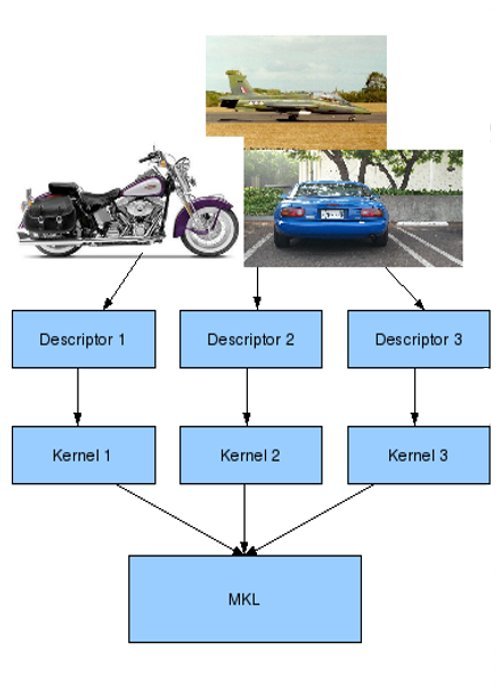}
\caption{Combining descriptors using MKL}
\end{figure}
\section{Optimal Combination of Kernels using Genetic Programming}\label{sec:obj}
For aiding object categorization in adverse conditions, many descriptors (and instance of feature representation) have been proposed in the computer vision literature. Suppose there are $n$ descriptors $d_1,d_2,...,d_n$ extracted from the $m$ images $I_1,I_2,...,I_m$. Using these $n$ descriptors, $n$ kernels $K_1,K_2,...,K_n$ of size $mXm$ are formed. Steps involved for producing the hybrid classifier using GP and SVM is described below.

\section{Steps Involved}
\begin{itemize}
\item Create a random population of kernel functions, represented as trees
\item Evaluate the fitness of each individual by building an SVM from the kernel tree and
   test it on the validation data
\item Select the fitter kernel trees as parents for recombination
\item Perform random crossover and mutation on the newly created offspring
\item Replace the old population with the offspring
\item Repeat Steps 2 to 5 until the population has converged
\item Build final SVM using the fittest kernel tree found from GP
\end{itemize}

\section{Parameters for GP}
The parameters for GP involves defining terminal set, function set and fitness function. Terminal set $= \{K_1, K_2, K_3, ..., K_n\}$, where $K_i$ is the kernel formed from any of the descriptors. Function set $= \{+, *\}$. Fitness function is the classification error of the particular chromosome on the training set. That is fitness value for each chromosome in this GP will be based on the accuracy of SVM with that chromosome (the non-linear kernel combination given to SVM). One alternative is to base the fitness on a cross-validation test (e.g. leave-one-out cross-validation) in order to give a better estimation of a kernel tree’s ability to produce a model that generalizes well to unseen data. 
\section{Results}\label{sec:exp}
The proposed idea was validated using real-world object datasets like Caltech-5 and Caltech-101. Caltech-5 contains five classes of objects cars, aeroplane, faces, leopards and bikes. Caltech-101 contains 101 categories of objects. Each category contains roughly from 30-100 images. Accuracy of the proposed method is compared to the best kernel $(K_1,K_2,...,K_n)$ and addition kernel which is $K_1+K_2+...+K_n$. All the experiments follow 1-Vs-1 SVM classification method. Binary classification mentioned in the following section is carried out by taking two classes at a time and accuracy is found on these two classes.
\section{Results on Caltech-5 dataset}
This section presents results on Caltech-5
%\footnote{\url{http://www.robots.ox.ac.uk/~vgg/data/data-cats.html}} 
using new MKL formulation and descriptors(csift, opponentsift, rgsift, sift, transformedcolorsift) provided from ColorDescriptor software. The experiments in this section are carried out using descriptors available from ColorDescriptor software
%\footnote{\url{http://staff.science.uva.nl/~ksande/research/colordescriptors/}}. 
Caltech-5 dataset contains images of airplanes, cars, faces, leopards and bikes. We have generated kernels on 5 descriptors provided using Gaussian kernel. This experimental procedure was repeated  10 times with different training-test data splits. It can be seen from Table 1 that the non-linear kernel method is giving better accuracy as compared to the best kernel and the addition of
kernels. Figure 6 shows plot of mean accuracy as number of iterations. Note that in all the iterations,  proposed non-linear kernel is better than other kernel combinations. Figure 7 shows plot of mean accuracy as number of binary classifier on Caltech5, totally 10 binary classification problem. Note that non-linear kernel combination gives higher accuracy than other kernel combinations in all binary classifications. Figure 8 shows non-linear kernel tree generated from GP which gives high accuracy than others on binary classification problem 9 in previous graph. In other binary classification problem, GP ends in selecting best kernel. Figure 9 10 11 shows some non-linear kernel tree generated from GP for Caltech5 dataset.

\begin{table}
\begin{tabular}{|l|l|l|}
\hline
Kernel & Caltech5 & Caltech101 \\
\hline
Addition of Kernels &  90.12$\pm$2.61  & 40.91$\pm$0.76\\ 
Best Kernel &  91.00$\pm$2.04 & 36.40$\pm$0.89 \\
Non-Linear Kernel &  94.76$\pm$1.71 & 42.71$\pm$1.48 \\
\hline
\end{tabular}
\caption{Percentage accuracy obtained using different kernel combinations}
\end{table}

\begin{figure}
\centering
\includegraphics[width=4in,height=4in]{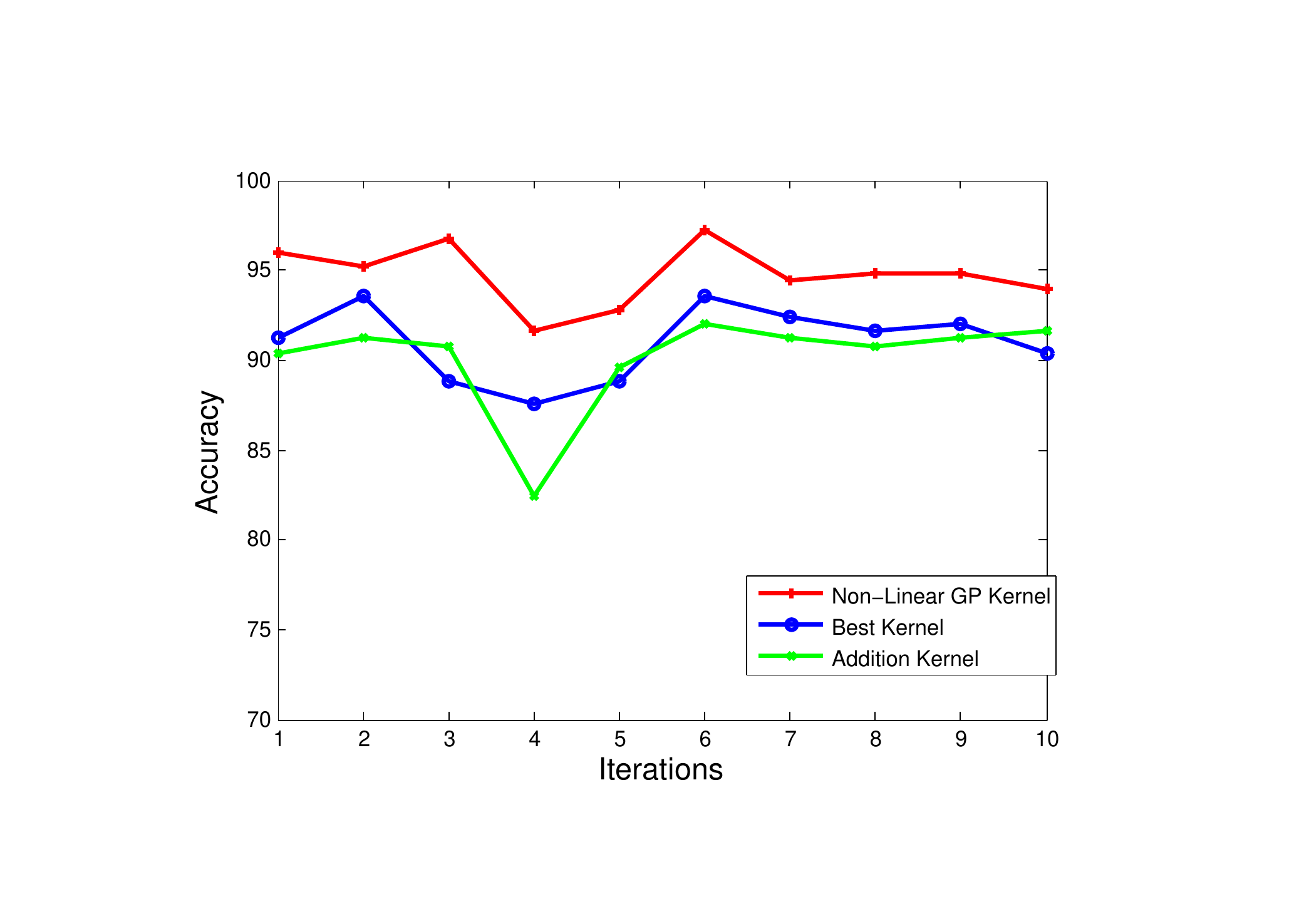}
\label{fig:nk_1}
\caption{Plot of mean accuracy as number of iterations.}
\end{figure}

\begin{figure}
\centering
\includegraphics[width=4in,height=4in]{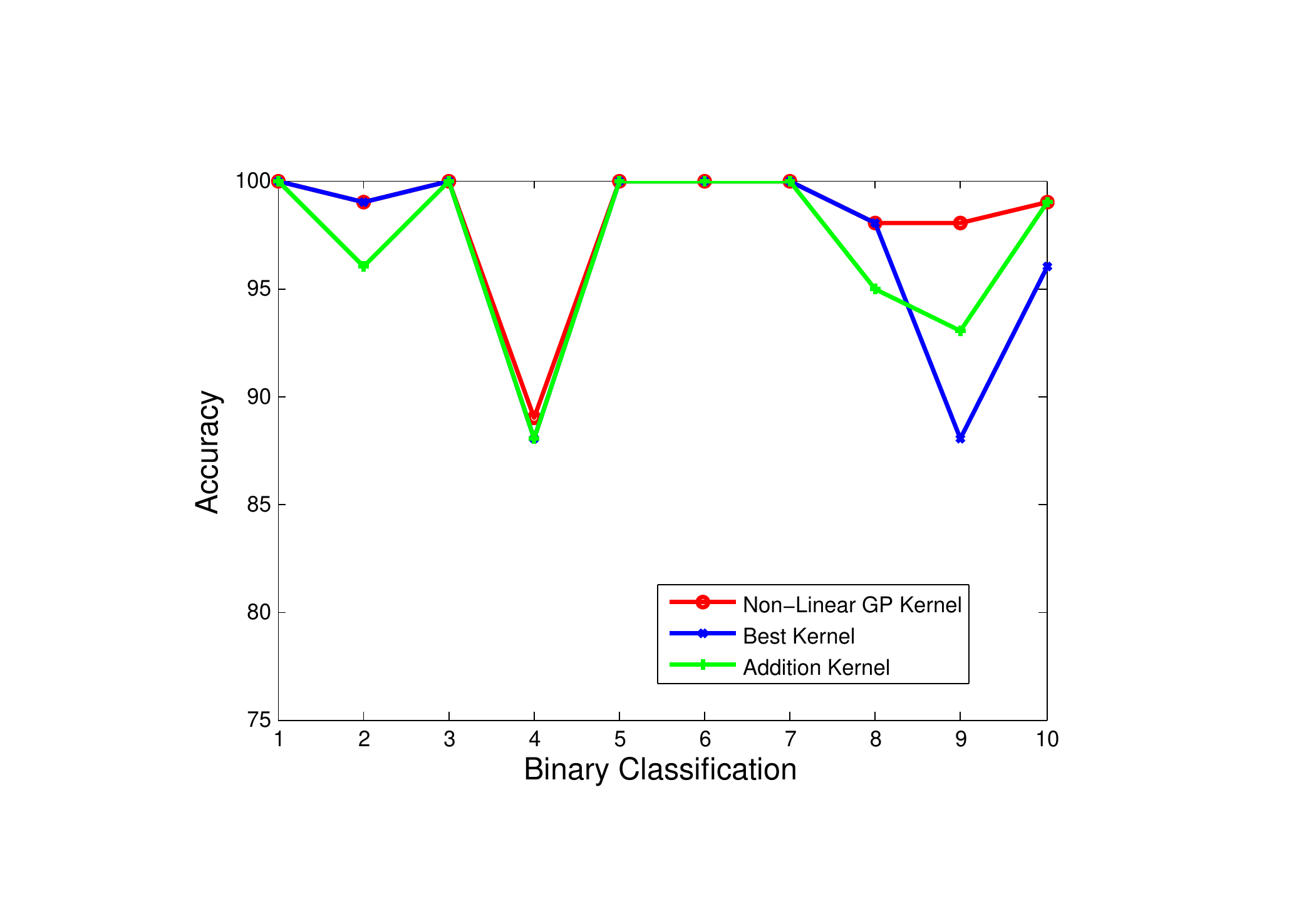}
\label{fig:nk_2}
\caption{Plot of mean accuracy as number of binary classifier on Caltech5.}
\end{figure}

\begin{figure}
\centering
\label{fig:nk_3}
\includegraphics[width=3.5in,height=3.5in]{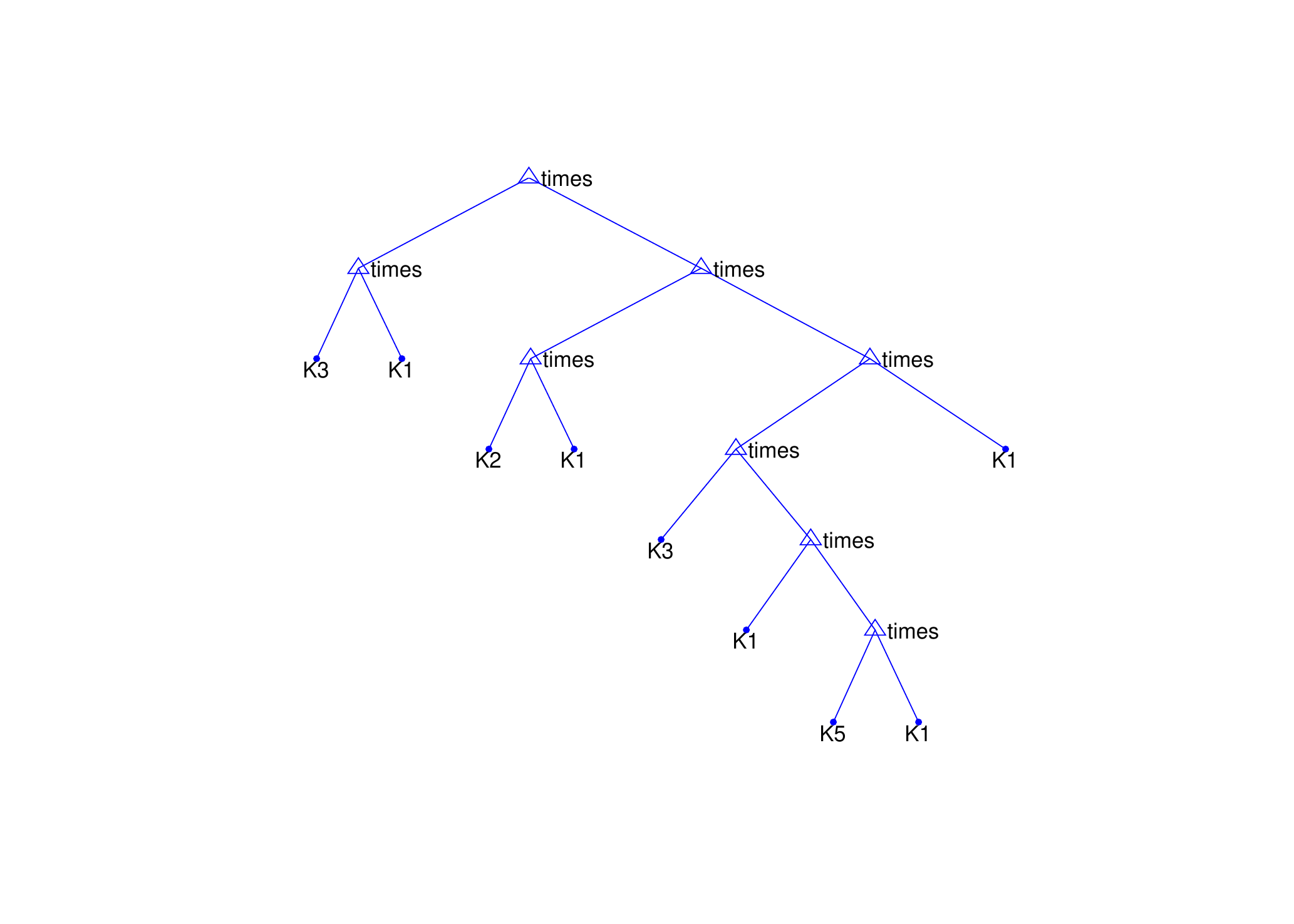}
\caption{Non-linear kernel tree generated from GP for Caltech5 dataset}
\end{figure}

\begin{figure}
\centering
\label{fig:nk_4}
\includegraphics[width=3.5in,height=3.5in]{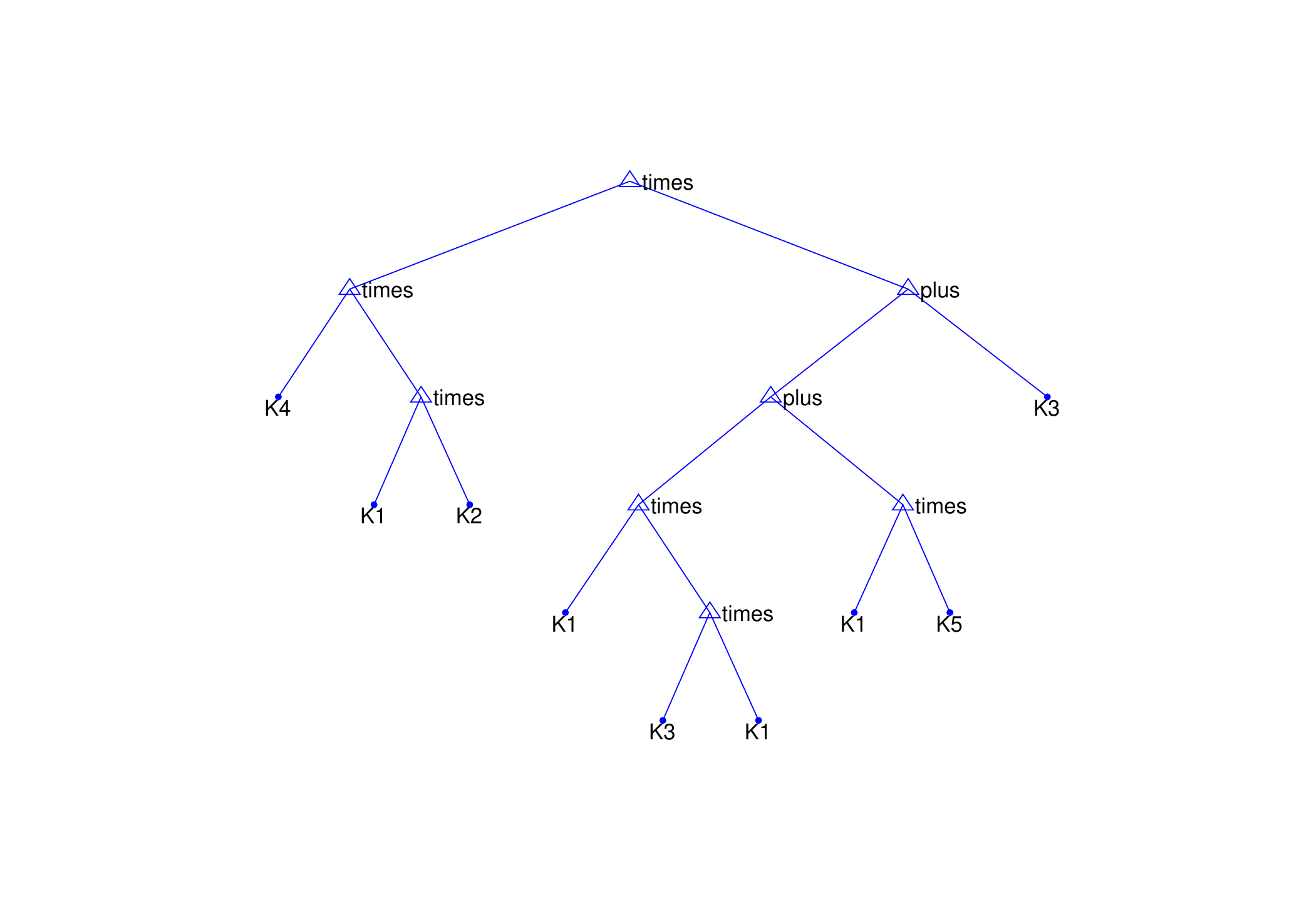}
\caption{Non-linear kernel tree generated from GP for Caltech5 dataset.}
\end{figure}

\begin{figure}
\centering
\label{fig:nk_5}
\includegraphics[width=3.5in,height=3.5in]{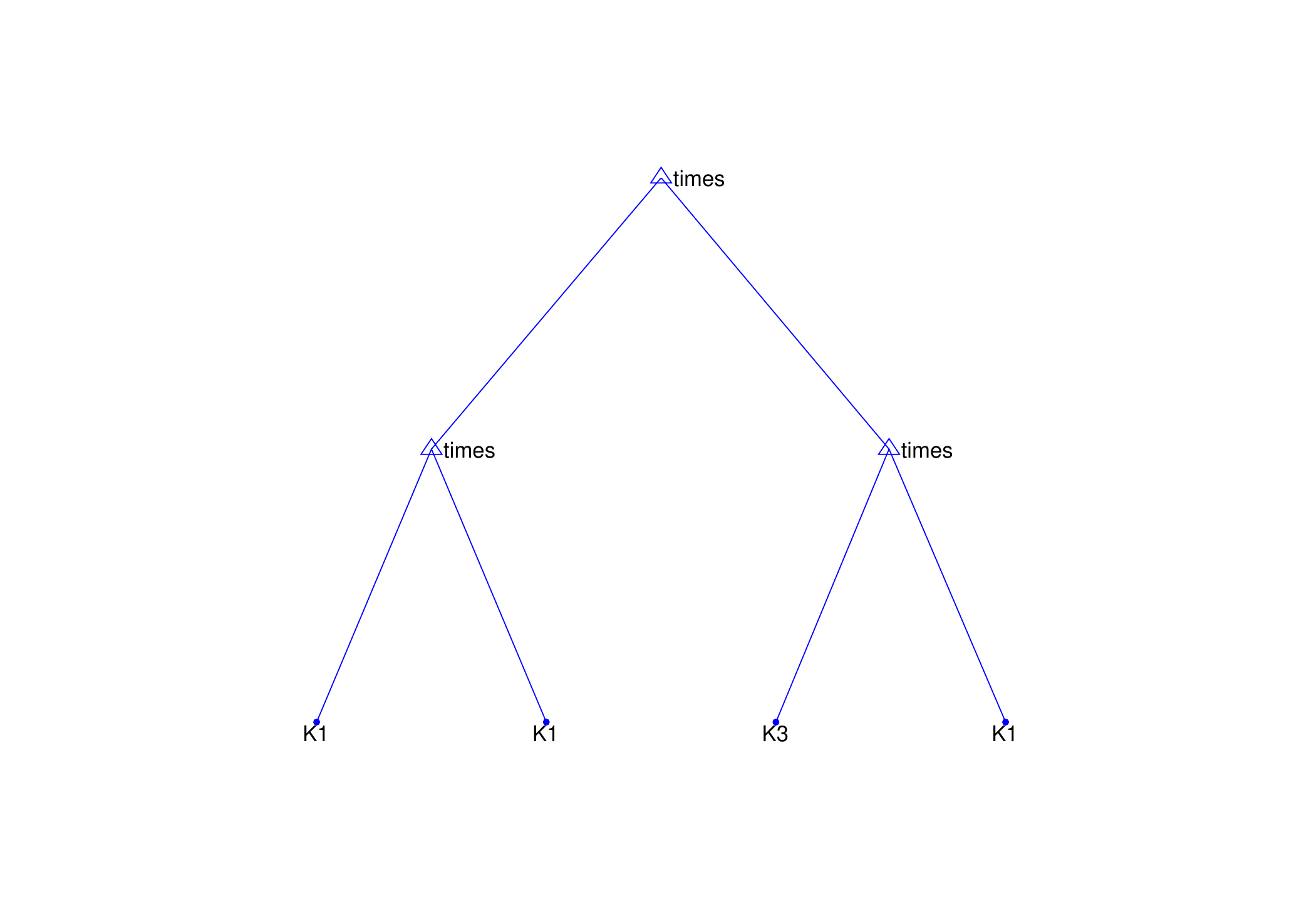}
\caption{Non-linear kernel tree generated from GP for Caltech5 dataset.}
\end{figure}

\begin{figure}
\centering
\label{fig:nk_6}
\includegraphics[width=3.5in,height=3.5in]{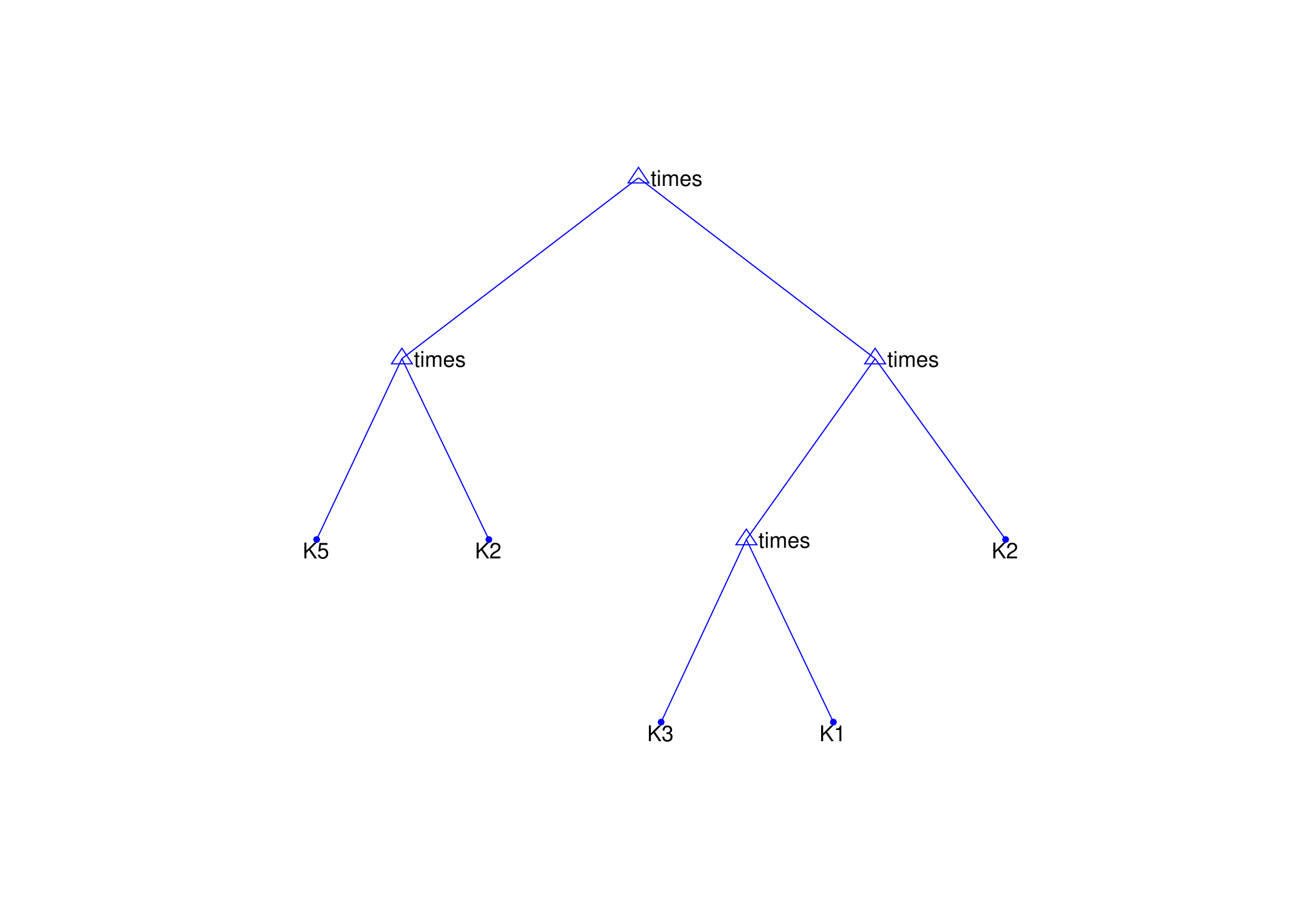}
\caption{Non-linear kernel tree generated from GP for Caltech5 dataset.}
\end{figure}
\section{Results on Caltech-101 dataset}
This section presents results on Caltech-101
%\footnote{\url{http://www.vision.caltech.edu/Image_Datasets/Caltech101/}} 
using new non-linear kernel combination and six kernels are taken from ucsd dataset %\footnote{\url{http://mkl.ucsd.edu/dataset/ucsd-mit-caltech-101-mkl-dataset}}.
 We have taken 30 images for each class, of which 15 are randomly taken as the training in which 5 are taken for validation data and the remaining as test data. This experimental procedure was repeated 5 times with different training-test data splits. It can be seen from Table 1 that the non-linear kernel method is giving better accuracy as compared to the best kernel and the addition of kernels. Figure 13 shows plot of mean accuracy as number of iterations in Caltech 101 dataset. Note that in all the iterations proposed non-linear kernel is the best. Figure 14 shows non-linear kernel tree generated from GP for Caltech101 dataset. This kernel tree is nothing but the addition kernel, which is generated for iteration 4 in Caltech101(see figure 13) where GP non-kernel and addition kernel give almost same accuracy. Figure 15 shows some non-linear kernel tree generated from GP for Caltech101 dataset.

\begin{figure}
\centering
\includegraphics[width=3.5in,height=3.5in]{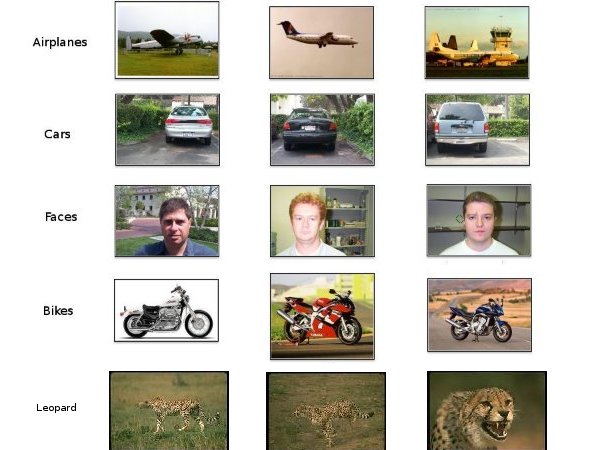}
\caption{Some Examples from Caltech-4}
\end{figure}

\begin{figure}
\centering
\includegraphics[width=4in,height=4in]{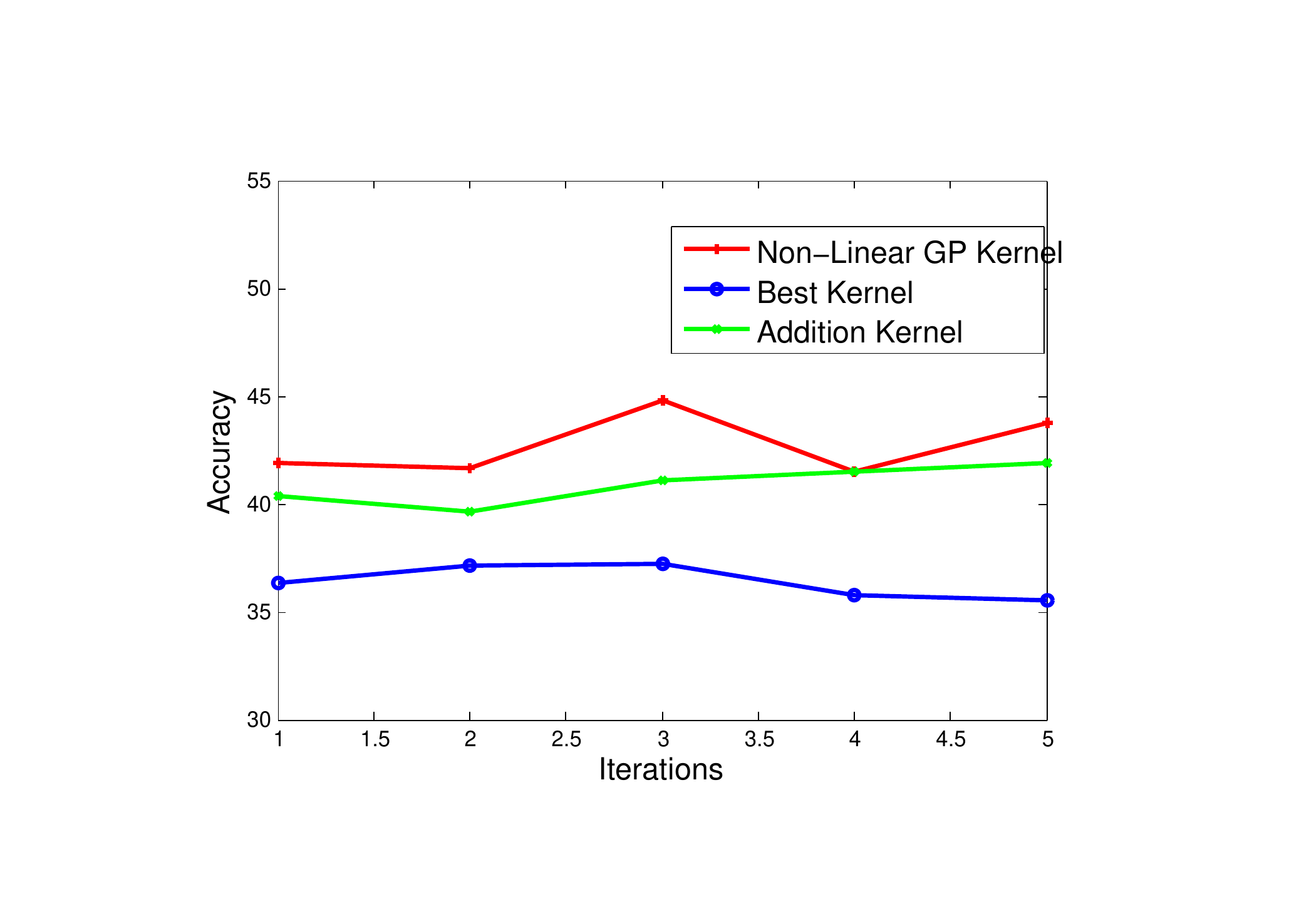}
\label{fig:mixed}
\caption{Plot of mean accuracy as number of iterations in Caltech 101 dataset}
\end{figure}

\begin{figure}
\label{fig:nk_7}
\centering
\includegraphics[width=3.5in,height=3.5in]{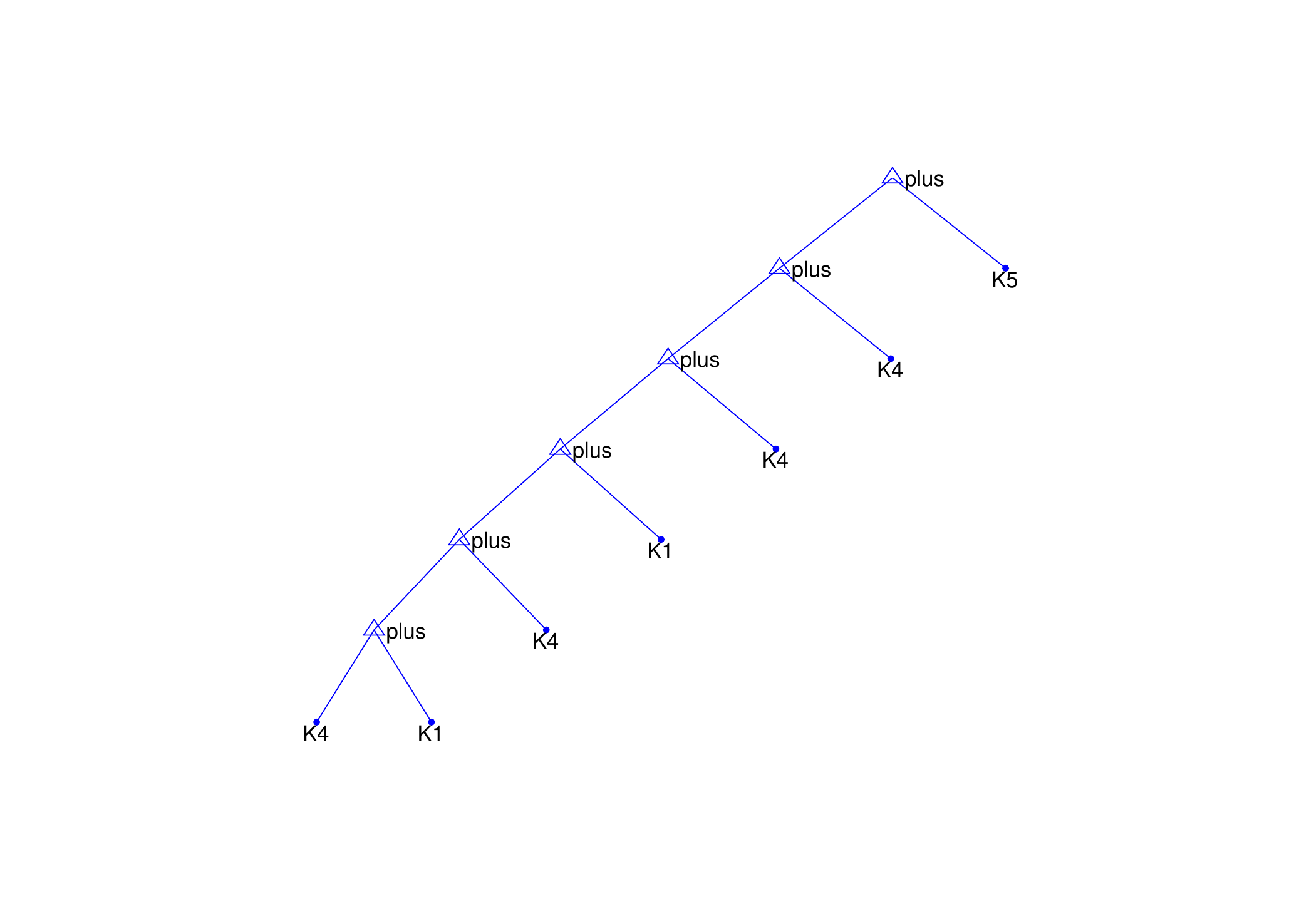}
\caption{Non-linear kernel tree generated from GP for Caltech101 dataset}
\end{figure}

\begin{figure}
\label{fig:nk_8}
\centering
\includegraphics[width=3.5in,height=3.5in]{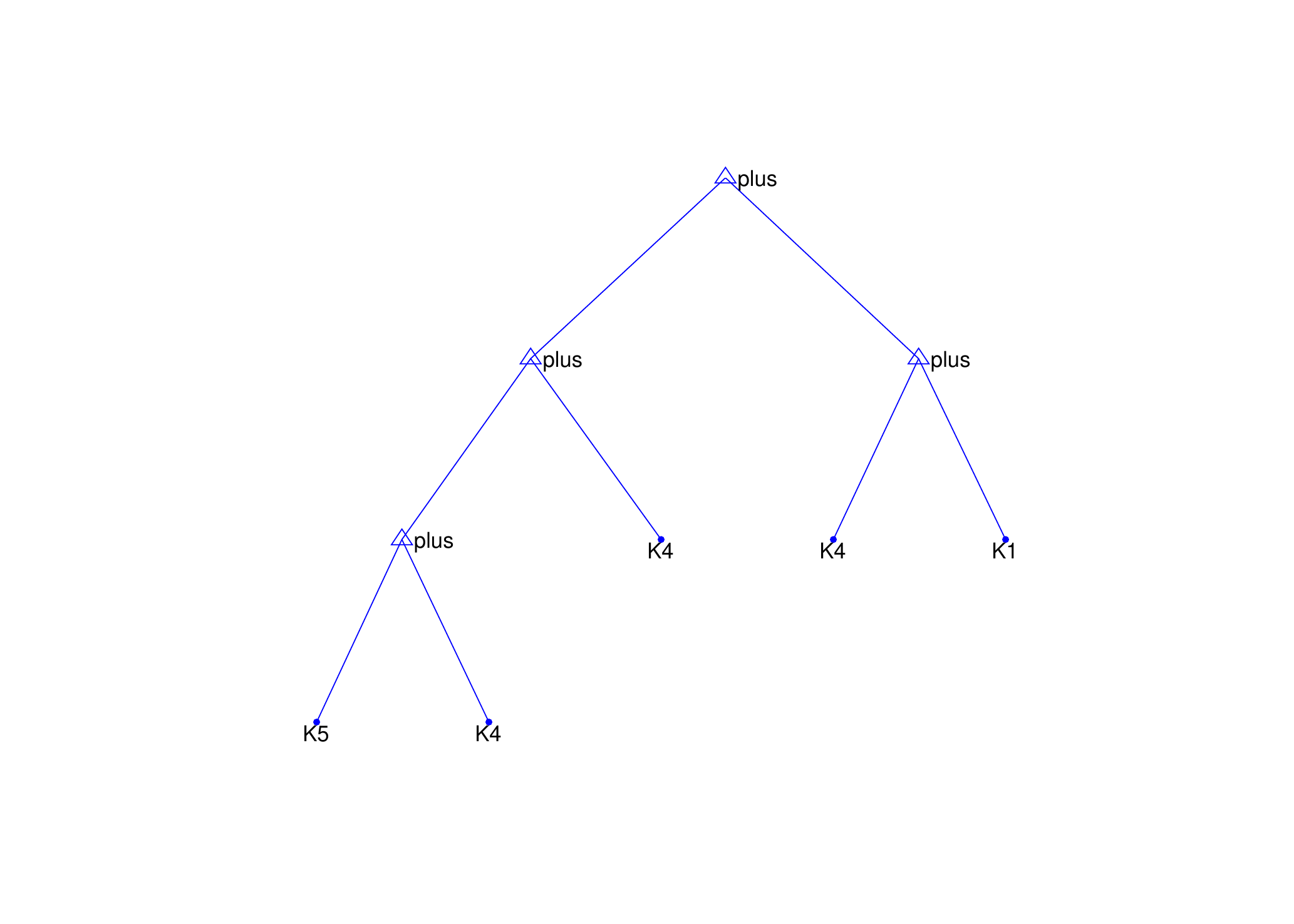}
\caption{Non-linear kernel tree generated from GP for Caltech101dataset}
\end{figure}
\section{Genetic Programming for Similar Images}\label{sec:prop}
Generated kernels from the previous sections are used for creating a demo
for similar images.  1530 images from Caltech-101 have been taken by us
and the kernels generated using the procedures discussed earlier.  A
matrix $M$ is generated which has dimension 1530 X 1530.  $M(i,j)$ refers to
the similarity between the ith and jth image.  Five descriptors are
generated for the images and therefore we have five kernels.  If the
addition of kernels is used then the kernels are combined linearly to find
M.  So $$M = K1 + K2 + K3 + K4 + K5$$.\\
We have used a non-linear combination of kernels found by using the GP. So, for example, M maybe
$$M = square(K1) + K1*K2 + K5$$ 
If a user clicks on the ith image, from the ith row in matrix M, the
minimum element is selected and the image corresponding to that is
selected as the most similar image.  The next minimum value gives the next
most similar image and so on.  These results are displayed and shown to
the user.  Figure 12 shows how the similar image demo works.  When an
image is clicked~\cite{dg5}, it shows the images most similar to that image.

%%%%%%%%%%%%%%%%%%%%%%%%%%%%%%%%%%%%%%%%%%%%%%%%%%%%%%%%%%%%%%%%%%%%%%
\section{Conclusions}
This paper proposed non-linear kernel combination using genetic programming. This eliminates the need for user to create non-linear kernel combination. Proposed framework is applied to Object Categorization. Experiments results shows that proposed framework for non-linear kernel combination using GP performance better than existing state-of-the-art kernel combinations.

%%%%%%%%%%%%%%%%%%%%%%%%%%%%%%%%%%%%%%%%%%%%%%%%%%%%%%%%%%%%%%%%%%%%%%
%%%%%%%%%%%%%%%%%%%%%%%%%%%%%%%%%%%%%%%%%%%%%%%%%%%%%%%%%%%%%%%%%%%%%%

% Bibliography or References
{\small
\bibliographystyle{ieee}
\bibliography{egbib}
}

\end{document}